# Model-Based AI planning and Execution Systems for Robotics

Or Wertheim and Ronen I. Brafman,
Department of Computer Science
Ben Gurion University of the Negev
orwert@post.bgu.ac.il, brafman@bgu.ac.il

*Abstract*—Model-based planning and execution systems offer a principled approach to building flexible autonomous robots that can perform diverse tasks by automatically combining a host of basic skills. This idea is almost as old as modern robotics [1]. Yet, while diverse general-purpose reasoning architectures have been proposed since, general-purpose systems that are integrated with modern robotic platforms have emerged only recently, starting with the influential ROSPlan system [2]. Since then, a growing number of model-based systems for robot task-level control have emerged. In this paper, we consider the diverse design choices and issues existing systems attempt to address, the different solutions proposed so far, and suggest avenues for future development.

*Index Terms*—AI-Enabled Robotics, Autonomous Agents, Integrated Planning and Control, Software Architecture for Robotic and Automation

## I. INTRODUCTION

MODEL-BASED planning and execution systems offer a principled approach to building flexible, autonomous robots that can perform diverse tasks by automatically scheduling a host of basic skills. This idea is almost as old as modern robotics [1], and diverse general-purpose reasoning architectures have been proposed in the past. However, truly useful systems that combine AI planners with modern robotic software platforms have emerged only recently, starting with the influential ROSPlan [2].

Model-based planning and execution systems for robotics (MPER) use a host of basic core capabilities, such as navigation, localization, mapping, manipulation, object-detection – which following [3] we refer to as *skills* – to solve diverse tasks automatically. To do this, they model the environment, the task, and the impact of each skill on the environment, and use automated planning algorithms [4], [5] to select which skill to execute and when. Because typical environments are non-deterministic and models may be somewhat inaccurate, these systems support state update based on sensing information, and modify their plan accordingly.

General-purpose MPERs can be configured by users to suit their own robot with its set of available skills. The user needs only provide the system with access to the skill code and a model of these skills in the system's modeling language. At runtime, the user provides, or the system detects or tracks the current environment state. Tasks can be provided to the MPER on the fly, which proceeds to activate the appropriate skills required to perform the specified task.

MPERs offer a powerful paradigm for software engineering of autonomous robotic systems. First, focus on providing core robot capabilities; this is what roboticists and signal-processing researchers have been working on for years, and what recent deep-learning and reinforcement-learning technology seems to excel at. An increasingly large body of code implementing such skills is available online. Second, describe your skill, environment, and task to the MPER. Ideally, given these two elements, the MPER can take over, controlling the system online towards successful task completion.

This approach to autonomous robot design addresses the two main software-engineering challenges in building general-purpose autonomous systems: *control* – how to use one's skill-code base to successfully perform each task; and *integration* – how to activate the actual skills and pass information between them. It also offers much greater accessibility to autonomous robot usage by lowering existing barriers: users can simply import a skill base (or buy it with their robot) and specify their tasks. Such accessibility can lead to many novel applications.

There are two major reasons for the recent emergence of general-purpose MPERs: (1) The widespread adoption of ROS as a standard programming framework for autonomous robots. (2) The increasing availability of software for diverse skills, such as navigation, localization, mapping, manipulation, object detection, etc., both within ROS and in general. Thus, ROS and its libraries now provide the needed code-base and platform for MPERs to thrive. For these reasons, these systems are often geared towards ROS platforms.

But there are also challenges to MPER deployment: the need to provide good formal skill models and the difficulty of capturing the intricacies of the real world inside such formal models, the need to possibly adapt them to particular robots and environments, and the requirement some MPERs impose on skill code – either by requiring additional integration code or by requiring that the code conform to some specific pattern. Moreover, MPERs require efficient inference algorithms to make good real-time decisions.

Building MPERs is a non-trivial software engineering project. They are not easy to design and implement because they combine different sophisticated pieces of software, yet must be general-purpose. Diverse design decisions must be made and then implemented, some of which are interdependent. They also require testing in diverse settings on real and simulated robotic platforms.

In recent years, several groups have implemented a number of MPERs. The goal of this paper is to examine the current state of MPERs, identify major ideas in current systems, and suggest areas in which they can be improved.

First, we explain why we believe MPERs provide a promising approach to the engineering of autonomous robots. Then, we describe the different components of these systems, the design choices they offer, and how these are addressed by existing systems. We conclude with a discussion of potential avenues for future improvements.

## II. WHY MODEL-BASED SYSTEMS

Model-based systems offer an understandable, controllable and transparent approach to the design of autonomous agents. They make our assumptions about skill codes explicit and potentially verifiable. The models can be read by users and can be improved, changed, and adapted to different settings. They also rely on optimization algorithms we understand and for which the user defines the optimization criteria.

MPERs also provide the most promising approach towards addressing regulatory concerns, e.g., as outlined in the EU's AI Act. First, skills must come with a model. Since skills are typically repetitive, circumscribed, and well-defined tasks, such as opening a door or pouring water, modeling them appropriately, using statistical information and programmer knowledge makes sense. Such a model can be viewed as a certificate. MPERs use optimization algorithms that operate using these skill models with clear objectives.

A model-based architecture can make system design clearer, easier and accessible to lay users, facilitating numerous new applications for autonomous robots. While not all systems provide this ease of use, in principle, users can download basic skills and their models and plug them into the system. Many more skills are likely to become available online in the future, making the range of potential applications larger and larger.

The benefits of MPERs are even clearer when we consider current practices. Today, task-level controllers are designed by writing complex scripts or state machines that perform them. These scripts must be rethought and revised upon any changes to the task, environment, or robot. Model-based methods do not require this constant manual revision of the controller. Instead, one revises the model, which is much easier and faster than revising the controller.

One major objection to the use of model-based systems in robotics is that models are difficult to provide. There are two aspects to this objection: the *subjective* element: much cognitive effort and expertise is required to formulate such models, and most programmers lack the knowledge and/or time required; the *objective* element: it is difficult or even impossible to capture all world nuances within a formal model for which good optimization algorithms exist.

We believe that the subjective difficulty is a non-issue. While we acknowledge that many users are not familiar with modeling languages, not all users need to provide a model. A good skill will be used by numerous users but requires just one model. If the payoff for providing such a model is clear, enough knowledgeable users will make the effort to provide it. We believe that much like open-source communities, open-model communities and contributors are likely to arise. In fact, skill models for current MPERs are not that complex. The language may be unfamiliar, but the information that must be provided is mostly common sense one that the programmers and the users are familiar with. With suitable interfaces, defining the model given this information is not such an onerous task. Moreover, various machine learning methods exist that can help automate this task. And perhaps obviating this whole issue, LLMs are showing a clear and increasing ability to generate such formal models from natural language descriptions [6] (also see [7], [8], [9], [10] and [11]).

The second point is a deeper one. It is indeed possible that existing models with their associated planning algorithms cannot capture all the intricacies of the world required for complex task execution. It is the role of current research to overcome this issue. However, it is possible to design good models, that compensate for such inaccuracies using various techniques such as replanning (a.k.a. model predictive control) and other mechanism that are already integrated into some systems, such as goal reasoning.

Of course, one could look to recent advances in AI, such as the use of LLMs for robot control or deep reinforcement learning, as potentially providing a more flexible, almost effortless solution. First, we are not there yet, and predicting how things will evolve seems foolish. Second, it seems that fundamentally, such systems do not come close to MPERs in terms of transparency, explainability, and controllability. Hence, it is not clear that we would like such systems to control autonomous robots in sensitive settings. MPERs allow us to combine the best of all worlds: use learning to learn repetitive, well-circumscribed skills, use LLMs to help create formal models, specify tasks using natural language, and help focus optimization algorithms on promising path, but keep the MPER as the master that selects what to do based on well-understood optimization algorithm and criteria.

## III. PLANNING AND EXECUTION SYSTEM REQUIREMENTS

In this section, we consider the diverse requirements from MPERs and some of the alternatives for fulfilling them. The key dimensions we consider are: *Modeling* – how the system models the robot's skills, task, and environment. *Decision Making* – how the system makes decisions and operates at run-time. *State Representation and Update* – how the system maintains and updates the current state. *Integration* – what integration effort this required to use this system on a robot. *Impact* – the technological maturity and impact of the system.

### 1. Modeling

The basic premise of an MPER is that it has access to a model of the robot's skills, the task, and the environment. This gives rise to the following questions:

1) What model is used and what aspects of the robot, environment, and task does it capture?
2) How is the model specified?

*a) Skill Models:* While all MPERs are, by definition, model-based, they differ in the models that they support. Each model has four fundamental components: the system state, how each skill changes this state, whether and how extrinsic changes or processes are modeled, and the task description.

In this paper, we use the term *world state* to refer to the state information exposed to the AI planner. The world state is often more abstract than the data needed to execute the robot's skill code. For example, the planner might use a discrete *kitchen* location, while a navigation skill will require *x, y coordinates*). We refer to this additional information required by the skills as the *execution state*, which, in some cases, is partially or fully included within the *world state*.

In all models, the *world state* is modeled as an assignment to a set of state variables that can have continuous or discrete domains. This is often referred to as a *factored* model (as opposed to a model in which the states are monolithic, structureless objects, e.g., $s1, s2, ....$)

When uncertainty is modeled, the world state is not necessarily known to the system. The two common ways of describing the knowledge of the system, typically called its *belief state*, are using a probability distribution over possible world states or using a set containing the possible world states. We refer to the former as probabilistic, or stochastic uncertainty and to the latter as non-deterministic uncertainty.[1] Here, too, a compact, implicit representation can be used. For example, a Bayesian network over the state variables can be used to describe the distribution over world states in stochastic models while logical formulas over the set of propositions can describe the set of possible world states in non-deterministic models.

In general, the models used correspond to various models used in AI planning, where the robot's skills are modeled as *actions*. The main features that differentiate them are:

- Type of state variables
- Certainty vs. uncertainty regarding the initial state
- Certainty vs. uncertainty regarding action effects
- Uncertainty form: stochastic vs. non-determinicity
- Instantaneous vs. durative actions, i.e., do we model the time it takes an action to execute and how it behaves during this time?
- Single skills at a time vs. multiple executing skills
- Fully observable state vs. partially observable state
- Accurate vs. noisy observations
- Goal-based vs. reward-based task description: In goal-based descriptions, the properties of desirable end states are described (e.g., *have-coffee*), and the robot attempts to reach such a state, possibly while minimizing cost, time, or other resources. In reward-based specification, a reward function describes the value of different states and the cost of different actions, and the agent attempts to maximize the total reward obtained during operation.

Existing systems use the following models, where *states* here refer to *world states*:

**Classical planning.** Propositional state variables are used. The initial state is assumed to be known, and skills are modeled as having a deterministic instantaneous effect on the state. A single skill can execute each time, and tasks are goal-based.

**Classical temporal planning.** Temporal models model skills as *durative* actions, i.e., actions with duration. Such actions have a richer structure, both in terms of the requirements for their successful execution, i.e., start, end, and overall conditions, and their effects, i.e., start and end effects. Concurrent skill application is allowed as long as the skills do not adversely interact. Current systems support goal-based classical temporal models.

**Contingent planning.** Contingent planning extends goal-based classical planning to model non-deterministic uncertainty, including non-deterministic effects, initial state, and sensing. It also models partial observability, i.e., where the robot cannot observe all aspects of the state. However, most contingent planners focus on partial observability and uncertainty over the initial state while assuming deterministic actions and accurate observations. Existing contingent models used in MPERs assume instantaneous actions, sequentially executed. Fully-observable non-deterministic planning (FOND) is a special class of contingent models with uncertainty only about action effects. We are not aware of a system supporting it.

**Stochastic models.** Models in which the uncertainty about skill effects, the initial state, and observations are described via probability distributions. In Markov decision processes (MDPs), we assume a known initial state, actions with stochastic effect, an accurate, fully observable state, and a reward-based task model. In partially observable MDPs (POMDPs), we assume an initial distribution over system states (an initial belief state), stochastic actions, partial observability with noisy sensing, and a reward-based task model. Action application is instantaneous and sequential.

**Multi-agent models.** Models that take into account the existence of multiple robots with potentially different states, where concurrent instantaneous actions can be applied by different agents.

*b) Model Specification:* Given that skills are modeled as actions within a planning formalism, models are typically specified using an action description language (ADL). STRIPs [1] was the first ADL, and it was developed specifically for robotics. Roughly speaking, ADLs can be divided into two categories: *declarative* (or *explicit*) and *generative* (or *implicit*.)

*Declarative* ADLs explicitly specify the transition induced by a state by describing the value of propositions in the next state or by providing the probability of each possible next state. They seek to be succinct and exploit the fact that typical actions change only a few variables. Hence, they focus on specifying the aspects of the world that change rather than spell out the next state or the distribution over it.

---

[1]In general, we will use the term *classical* to describe non-stochastic models. In the literature, *classical* often refers to deterministic models, but it is often also used to differentiate models with non-deterministic uncertainty and stochastic models.

*Generative* ADLs provide a way to simulate the transition. Although one can simulate a transition given a declarative model, to efficiently specify dynamics that are difficult to describe or simulate using a declarative model. On the other hand, retrieving the specific transition probabilities from a generative model may be difficult.

**Planning Domain Definition Language (PDDL).** PDDL is the main explicit ADL for classical planning. Originating from the 1998 International Planning Competition (IPC), PDDL is an advanced logic-like language that can model both classical planning and temporal planning with concurrent durative actions. PDDL Domain files define general domain features, such as object types and the action schema. Specific problem aspects are contained in a separate problem file. Almost all classical planners today support PDDL as their input format.

**Relational Dynamic Influence Diagram Language (RDDL)** Introduced for the 2011 IPC's uncertainty track, RDDL is an ADL for specifying MDPs and POMDP. RDDL uses dynamic Bayesian networks (DBNs) to specify the impact of actions on state variables. It can model partial observability and allows for concurrent (instantaneous) actions. Distributions can be specified using a rich language that supports various arithmetic and logical operations. However, it does not support iteration and complex data structure. Hence, RDDL has both declarative and generative features. While highly expressive, modeling more complex probabilistic processes may require large and complex network structures. Numerous planning algorithms that supported the IPC probabilistic and Reinforcement Learning track have RDDL interfaces.

**Skill Documentation Language (SDL)** SDL was designed to model POMDPs. Like other ADLs, the user describes the state variables. However, the actions (the transition and observation functions) can be described using enhanced C++ code. It can be viewed as employing a probabilistic programming language (PPL) to describe the model, in the spirit of earlier PPL-based specification formats [12]. The powerful constructs available within a PPL make it easier to specify more complex models and may be easier for programmers to use. However, because it is less declarative, it may be harder to understand and analyze. For example, existing techniques for generating heuristic functions typically require a simple declarative format.

## 2. Decision Making

Given a model, decisions must be made at run-time. While a planner is the key component, it is often part of a larger mechanism. How does this mechanism work? How does the planner represent its state at run-time, and what mechanism is used to update it?

By definition, MPERs rely on planning algorithms for decision making. However, there are different issues that arise and different techniques that complement the use of a planner. These include the use of replanning (a.k.a. model predictive control) to handle inaccuracies in the model, a common source of which is the use of deterministic skill models, the use of goal-reasoning mechanisms to drive the planning process, and the use of pre-specified plans and goal decomposition methods. We discuss them below.

**Using a Planner**. The robot acts online, using a planner to decide how to act. However, decisions can be made online or offline. If made offline, an entire plan (i.e., an action sequence) or a policy (e.g., a policy tree) is generated, which is then used to select which skill to apply and when. In that case, the robot only needs to look up the next action in the sequence or the tree and execute it. If decisions are made online, the agent invokes the planner in each step in order to compute the next action.

Offline plan (as opposed to policy) generation is rarely a true option because this assumes a deterministic world. Robots rarely operate in a deterministic environment. Instead, MPERs that utilize planners that generate sequential plans employ the idea of replanning, (a.k.a. model predictive control). Following each action, the system updates its state, and if the observation is unexpected, it plans again from its updated state. The advantage of this approach is that one needs to represent within the system a world state rather than a much more complex belief state. Planning in these models is typically simpler and more efficient. The disadvantage is that the planner does not reason about uncertainty. It selects actions based on strong, often unrealistic assumptions about the state of the world and the effects of actions, does not reason about tradeoffs between potential outcomes, and has no motivation to seek information actively.

Offline policy generation makes more sense, as it takes into account the different possible observations. However, in many cases, this is not feasible computationally. Offline POMDP solvers have difficulty scaling up to models with large state spaces, (e.g., a world with many manipulable objects) and large or continuous observation spaces. Even in MDPs, the focus has shifted to online solvers, which scale up better.

Online planners select a single action at a time, without committing to future actions. But their choice is informed by the different possible futures, and in the stochastic case, also their likelihood and their value. Online planning is more efficient since it does not need to compute an entire policy tree and, therefore, scales to larger problem instances. Many online algorithms are any-time, i.e., can always return some action, with the quality of their choice improving given more planning time. While, in principle, many online algorithms have convergence guarantees in the limit, in practice, their choices are sub-optimal given limited time.

**Goal reasoning** [13]. Goal reasoning encompasses an agent's capability to determine its current objectives autonomously. The goal reasoners assess the current situation, the history, and the task specification and divides a complex task into smaller tasks, called goals. It selects which goal to pursue next and determines whether the current goal should be abandoned and replaced. The use of goal reasoning allows using simpler planners (such as classical planners) in a dynamic, uncertain environment, and results in a potentially better-informed behavior than using simple replanning. Goal reasoning is reminiscent of the principles behind hierarchical planning [14] or Options [15], as both break a complex task into sub-components.

**Pre-generated Plans and Decompositions**. Sometimes, the programmer may wish to specify, ahead of time, certain strategies for achieving some set of goals. This may be to ensure safety or other constraints, or simply to enhance efficiency in tasks or sub-tasks that must be achieved frequently. This capability is related to goal-reasoning and the use of hierarchical models, and is supported by some systems.

**Sensing and Monitoring**. In non-deterministic domains, agents must use their sensors to detect the current world state and respond accordingly [16]. *Sensing* involves updating the world state based on sensor data without directly intending to affect the robot's current behavior. It can be done in different ways: *synchronous sensing* is when the world state is updated immediately after a skill terminates execution. This could either be in the form of the return value of the skill or by applying a specific sensing skill. *Asynchronous sensing* occurs when sensing information is obtained continuously as part of an executing skill or as a background process that actively updates the state. We use the term *passive sensing* to refer to the latter, i.e., to background sensing activity that is not modeled as a skill, and hence is not explicitly scheduled by the planner. For example, the robot may continuously sense and update its position throughout its lifetime without an explicit localization skill being scheduled by the planner. Passive sensing is, by definition, asynchronous and requires explicit support by the MPER, often in the form of an API enabling direct update to the world state (see below).

*Monitoring* is sensing with an associated intention to respond that occurs outside a skill. The response could be to replan, to interrupt the current action and/or to invoke some other response. In general, this term is used to refer to monitoring activity that can occur within a skill (e.g., localization to ensure reaching a target position) or monitoring that is taken care of by the planner (e.g., applying an explicit sensing action that checks whether a *pick* action was successful.) However, we use this term to refer to an explicit mechanism that the MPER provides for sensing and/or performing corrective behavior that is not explicitly invoked by the planner. For example, the MPER may support code that monitors an executing action to determine if it is stuck, failing, or taking too long to execute, in which case, the MPER can respond by halting execution and replanning. Or, the MPER may support code that monitors the state for some unsafe conditions and will trigger, in response, some corrective action.

### 3. State Representation and Update

Earlier, we discussed the static skill model representation used by the planner. Another key issue for MPERs is what world-state model is maintained and how it is updated. This greatly impacts, and is related to the manner in which information is passed from the planner to the skills and back, how skills pass information to each other, and how external entities, such as other robots or users can interact with the system. The key questions are what information is stored and where, and what access to it do different entities have. For example, can skills pass low-level data that is not maintained by the planner to each other? Can the planner provide additional information that can assist the low-level code make better choices. For example, the arm manipulator could place an object differently if it knows the planner will shortly request that it be picked up.

It is clear that the planner's *skills model* and the *world state* must be maintained by the system. However, the *skills model* is often abstract and operating the robot often requires lower-level data, the *execution state*. For example, the planner may use as object-location-values the rooms they are in, but the skill code would typically need precise coordinates. Some MPERs make no distinction between these elements maintaining all together, while some differentiate between the *world state* and the *execution state*.

More important is the approach to update. Some systems maintain a clear abstraction barrier – skills do their work and return values and the system uses this to modify the *world state*. Other systems expose an API to the *world state*, allowing external code to modify it directly. This is more flexible since it enables other entities, such as other robots or users, to update the state directly. In the first case, one must centrally specify all possible return values and their impact. In the latter case, this information could potentially be handled by multiple processes. However, this also implies that the system does not control the state directly, and that users must write explicit update code. We note that most systems that attempt to support multi-robot applications use the latter method. A central storage (possibly external to the robots, as supported by ROS) is made accessible to all agents via an API, implementing a shared state.

### 4. Integration

A key issue in evaluating MPERs as a suitable SE tool is understanding and assessing the effort in the process of integrating or adapting skill code with the MPER. Several factors need consideration. What steps are required for integration, and how difficult is each step and the entire integration method? Does the system facilitate this process through supportive actions, or perhaps by fully automating some of the steps, or through a user friendly API? How easy is it to write the abstract model for decision-making? Are there tools to help the specification process? What are the challenges involved in linking this model with low-level code? In particular, can low-level code be imported and treated as a black box, or must it comply with certain design-patterns? Finally, what is the learning curve associated with using the MPER? These questions are crucial for understanding the time and effort required from the robot's engineer.

*5. Impact*

Ultimately, the goal of MPER designers is for their software to be used to build robots. Has this hope materialized? What tool and documentation is offered for training users?

## IV. Systems

*A. ROSPlan*

ROSPlan pioneered the integration of planning engines with skill code in the context of ROS, and is thus the first modern MPER. It also has supports the widest range of planners and models and extensions.

*1. Modeling:* **Supported Planning Types**: ROSPlan supports classical, temporal, and contingent planning with instantaneous or durative actions, including concurrent actions.

**ADLs**: Primarily, ROSPlan uses PDDL (Planning Domain Definition Language). There is experimental support for RDDL (Relational Dynamic Influence Diagram Language), PPDDL (Probabilistic PDDL) and HDDL (Hierarchical Domain Definition Language), which was the standard input language for hierarchical planning in the 2020 IPC.

**Probabilistic Environments**: For environments with probabilistic effects, ROSPlan has an RDDL extension. An approximate deterministic model combined with replanning is also an option.

**Limitations**: ROSPlan struggles with partial observability and noisy sensing. Relying on approximated models in these conditions can lead to robots exhibiting repetitive behaviors due to conflicting noisy observations. Its RDDL extension can derive plans for (PO)MDPs (Partially Observable Markov Decision Processes), but it is limited to domains in which the robot can perfectly sense any state variable whose value is a stochastic outcome of a skill, which is not the case in many such domains.

*2. Decision Making:* **Planners.** ROSPlan supports a wide range of planners that users can invoke depending on the action description language (ADL) they use. Both online and offline planners are supported. Replanning is employed when a deterministic model approximates uncertain worlds, based on the sensing and monitoring of the robot's execution.

**Goal Reasoning.** Not supported.

**Pre-generated Plans and Decompositions.** ROSPlan enables the creation of finite state machines (FSM), which can be utilized to integrate robot skills for accomplishing subtasks. Through ROSPlan's Action Interfaces, these FSMs can be defined as a single PDDL action that encapsulates their overall effect and conditions. Additionally, ROSPlan offers experimental support for HDDL, allowing users to incorporate domain-specific knowledge through the hierarchical decomposition of the problem.

**Monitoring and Sensing.** ROSPlan endorses *active sensing*, where the planner calls skills to gather information and update its knowledge base (KB). Its sensing interface [17] supports *passive* and *asynchronous sensing* through a background process that continuously updates the system state by analyzing public data published in the robot's framework, such as ROS topics. The sensing interface also allows periodic calls to ROS services to analyze and update the current state in the KB.

Users have the flexibility to prompt replanning either after or during a skill's execution, indicating that ROSPlan caters to both *synchronous monitoring* and *asynchronous monitoring*. However, the implementation of monitoring is the user's responsibility, with ROSPlan providing the necessary APIs.

*3. State Representation and Update:* The domain and problem files, along with the current state, are stored in PDDL format within the ROSPlan Knowledge Base. APIs are available to query and update the *skill model* and *state model* described within these files, as well as the plan generated by the solver. These APIs are implemented as ROS services, making them accessible from different machines. This provides significant flexibility for human users or other agents to update the Knowledge Base (KB), altering goals or the robot's state to meet specific needs (e.g., task changes or knowledge updates from another agent). The APIs are also available for robot skills to provide valuable information to the planner or to query the current plan to consider future skill activations. This approach offers great flexibility but comes at the cost of clear encapsulation. ROSPlan maintains a deterministic *world state*. Nevertheless, within this state, it is possible to assign state variables the value *unknown*, providing a limited capability to describe belief states. Specifically, this is limited to non-stochastic belief states (i.e., a set of possible world states) where the values of state variables are not correlated.

ROSPlan only manages the *world state*. Transformations to the *execution state* (e.g., transforming action parameters to code parameters) are outside the scope of the system and must be handled by user code.

*4. Integration:* ROSPlan users should supply their ROS skill code along with a PDDL description of the task, environment, and skill behavior. To integrate skill code with the system, users must instantiate the ROSPlan *Action Interface* for each skill.

Users should extend a template action class by writing the skill code activation based on the action parameters. The *Action Interface* handles subscription and publication to ROS topics that dispatch the action, fetches operator details, checks conditions before execution, and updates the knowledge base with the action's effects.

If users want to implement custom behavior (e.g., if the action sometimes fails and the effects should not be updated), they must write the Action Interface from scratch.

For sensing the environment, ROSPlan's *Sensing Interface* offers a configuration file to define simple, single-line Python translations from a topic's ROSMessage to a predicate value update. For more complex translations, users can write a Python function that the interface automatically calls. However, this function is limited to handling a single predicate. Additionally, the interface can periodically invoke a sensing ROS service based on user code for more complex updates.

The *Action* and *Sensing Interfaces* significantly improve ROSPlan's ease of use. However, the necessity to write code within ROSPlan's code base when the configuration files are insufficient is cumbersome.

Activating planning or replanning and dispatching the plan is the user's responsibility, requiring them to write their own code that wraps ROSPlan's APIs.

In addition to the various ROS interfaces available for monitoring and debugging (e.g., rqt, command-line tools, logging), ROSPlan offers a GUI. This graphical user interface allows users to load and clearly display Action Interfaces and observe which actions are activated at each point.

*5. Impact:* ROSPlan offers detailed documentation [18] and tutorials for using the system. Its impact on robotic research is demonstrated by the number of projects that utilize the system (see [19], [20], [21], [22], [23], [24], and more). Nevertheless, ROSPlan supports ROS1, which is becoming obsolete and to which it is tightly bound. Although ROSPlan remains a cornerstone from which MPER developers can learn, its strong reliance on ROS1 poses a significant challenge for maintaining its influence in robotics, especially as more systems migrate to ROS2.

**In conclusion**, ROSPlan offers a rich set of capabilities for planning with multiple planners, executing generated plans, and an easy way for various stakeholders (e.g., other agents, human users, skills code) to modify the state and goals. However, it lacks robust support for partial observability and noisy sensing. It primarily supports the PDDL model, which has some limitations in expressive power (See [25]). Yet, its sensing and action interfaces greatly improve system usability.

### B. CLIPS Executive (CX) [26]–[30]

The CLIPS Executive (CX) system employs a predefined high-level plan, structured as a goal tree. Goals at the leaf nodes of this tree are achieved using a PDDL solver, which generates plans that invoke specific robot skills. System information is stored in NASA's CLIPS reactive expert system engine. The engine refines the goals and assesses their feasibility based on updates received from the skills about the *world state*. It actively modifies the selected goals, using the goal tree as a basis for changes in goal selection. The effectiveness of this system has been demonstrated through its success in winning robot competitions (2018 RoboCup German Open and PExC).

**CLIPS.** CLIPS is a rule-based system developed by NASA as an expert system. It consists of three main components: a fact list, a knowledge base, and an inference engine. The fact list is a collection of facts that the system uses. The knowledge base contains procedures and rules. Procedures, written in C++, can modify the fact list or execute robot code. Rules include a list of preconditions on the fact list; if these conditions are met, the associated procedure is added to the agenda. The inference engine constantly evaluates the rules and determines which procedures to execute. Multiple rules can run simultaneously, and rules are executed until the system reaches a stable state, at which point the agenda becomes empty.

We now examine the different aspects of the system.

*1. Modeling:* CX supports the PDDL action description language for both classical and temporal planning.

*2. Decision Making:* CX does not support probabilistic skill outcomes, partial observability, or noisy sensing. The system assumes deterministic outcomes for actions and complete knowledge of the state, which can be a major disadvantage in real-world scenarios where uncertainty and incomplete information are prevalent. To compensate for this, it relies on the ability to support more flexible sensing, monitoring, and goal reasoning, empowered by integration with CLIPS.

**Planners.** CX utilizes PDDL to define planning problems and supports two robotic frameworks, each with different planners. For the Fawkes [31] framework, CX is compatible with the FF [32] and FastDownward [33] planners. In the context of ROS2 (see [26]), it employs the Plansys2 [34] MPER and can utilize both the POPF [35] and the TFD [36] planners.

**Goal reasoning.** CX's goal tree serves as a predefined hierarchical high-level plan but goes further by continuously evaluating and reasoning about the goals using the CLIPS engine. CX has a goal refinement mechanism, an ongoing assessment of the goals' status, allowing it to adapt to changing conditions and priorities, enhancing its flexibility and responsiveness. External events, such as a skill updating the *world state*, may trigger rules and procedures in CX that change the goals' lifecycle state.

The lifecycle of a goal in CX begins with formulation, where the goal is defined within the goal tree. Next, it can be selected for execution. If the goal is complex, it is expanded into sub-goals; otherwise, a solver is activated. Once expanded, the goal is committed to a plan generated by the solver, resources are allocated, and actions are dispatched.

After execution, the goal reaches the finished stage, where the result is determined (e.g., succeeded or failed). It then moves to the evaluated state, where the *world state* is updated by user-defined criteria. Finally, the goal is retracted, and resources are released. This lifecycle, combined with continuous assessment, enables CX to respond effectively to dynamic environments.

CX's complex goals include sub-goals, with control structures such as "run-all," meaning success if all sub-goals succeed, or "try-all," meaning success if at least one sub-goal succeeds. This adds complexity and adaptability. Leaf goals represent PDDL objectives to achieve or conditions to maintain.

**Pre-generated Plans and Decompositions** CX allows simple goals to be achieved either through a PDDL solver or by activating a predefined sequence of skills. This feature is convenient for defining macros that achieve a combined effect but is less powerful than FSMs supported by other systems in this context.

**Monitoring and Sensing.** CX can monitor actions, plans, and goals through a sensing and monitoring mechanism that supports both *synchronous* and *asynchronous sensing*. Robot skills, such as object detection or background pro-cesses, can update the *execution state* at any time by adding facts in CLIPS, enabling CX to support both synchronous and asynchronous sensing. The updated *execution state* is then translated into the planner's *world state*, also stored as facts in CLIPS, and utilized for planning. The system continuously monitors the validity of actions, plans, and goals, and execution monitoring may trigger replanning if necessary. CX's *action monitoring* verifies the preconditions of actions before execution using its knowledge base. For *plan monitoring*, it incorporates corrective behavior in response to exogenous events. Additionally, goals are evaluated, and their outcomes alter the fact list, which may result in the rejection of other goals and ultimately change the selected goal.

*3. State Representation and Update:* In CX, the Goal tree the PDDL domain and problem files, and the *execution state* are stored in CLIPS using an extended PDDL format. Specifically, CLIPS maintains several models: The **Domain model** is similar to the PDDL domain file and contains operator definitions and object types. The **Planner model** is akin to the PDDL problem file and includes facts about predicate and object instances (*world state*). The **Execution model** extends the PDDL domain with information about sensed predicates. For example, a predicate can be marked as *sensed*, in which case any skill that modifies it is immediately followed by a sensing action that evaluates it. The **World model** extends the Planner model with all known information, including facts that are irrelevant for planning but necessary during execution, such as precise positions and information about other robots (*execution state*). This model supports rich data formats, including lists, numbers, symbols, and strings. Together with action feedback, the world model provides the primary interface for ingesting information by the robot into the execution.

These models collectively store both high-level and low-level data. The system state stored in CLIPS encompasses the knowledge in the Goal Tree, the current life cycle state of each goal, the Planner domain, and the World model.

**Interactions.** CX data saved in CLIPS is accessible to both skills and external entities (e.g., human users), allowing them to influence the system. The rich state stored in CLIPS, which includes both the world state and execution state, facilitates information exchange between skills. A skill can save data in the execution state, making it accessible to other skills, and can update the world state used by planners for abstract decision-making information. Additionally, skills can access the generated plans.

While both CX and ROSPlan allow direct updates to the database, prioritizing usability over encapsulation, ROSPlan's updates are mediated through ROS services that interact with the database, whereas CX users directly access the CLIPS database, further emphasizing flexibility at the expense of encapsulation and data safety.

**Multi Agent support**. CX provides unique support for multi-robot scenarios in which each robot plans for itself, but the robots communicate to resolve conflicts and coordinate. They share parts of each robot's model using database replicas, so each robot has information about the state of the others. They also use a resource-locking mechanism, so a robot can only commit to a goal if it has acquired all the needed resources. CX also offers an option by which a single central agent controls all the robots.

*4. Integration:* CX users are required to provide a configuration file detailing specific system properties, the PDDL domain file, and a goal tree, and they must implement each robot skill (also referred to as a feature) using a base class. Updates to CLIPS facts, which are integrated within the skill code, must be defined by user code. Additionally, users must define a PDDL goal for each goal-tree leaf. The CLIPS rules must also be defined to allow for monitoring and goal reasoning. The integration effort required by CX users appears to be substantial. While the system supports complex scenarios effectively, this capability requires non-trivial programmer effort.

*5. Impact*

CX has been successfully implemented and tested across various scenarios. It boasts several conceptual and engineering advantages, as evidenced by its success in winning robotic competitions, including the 2018 RoboCup German Open and PExC.

Currently, most information about the system is available primarily through research papers published by the development group [30] [27] [37] [26] [38]). Facilitating use by other potential users calls for more extended GitHub pages (See [39] and [40]) and additional documentation and tutorials.

*C. Plansys2*

PlanSys2 incorporates concepts similar to those of ROS-Plan, but it is designed for ROS2. While ROSPlan supports a broader range of planners and ADLs and includes Sensing and Action interfaces that simplify integration, PlanSys2 offers more accessible and extensive programming interfaces. It also features a more straightforward and structured approach to system activation. Crucially, it supports ROS2, which is becoming the de facto standard for robotics.

The design of PlanSys2 includes Domain and Problem experts, which manage information about the PDDL domain (types, predicates, functions, and actions) and problem (instances, predicate values, function values, and goals) respectively. A planner component utilizes this information to generate plans through an AI solver, and an Executor is responsible for implementing these plans by executing the robot's code. Each robot skill is integrated using an ActionExecutorClient node.

Plansys2 offers several engineering advantages, one of the most significant being its ability to automatically translate a temporal plan containing concurrent actions into a behavior tree. This feature enables efficient concurrent execution of the plan.

We now describe the different aspects of the system.

*1. Modeling:* **Supported Planning Types:** Plansys2 supports the PDDL action description language for both classical and temporal planning.

*2. Decision Making:* **Planners.** By default, Plansys2 uses the *POPF* [35] planner, and can accommodate the *Temporal Fast Downward* [41] planner. Both are offline, temporal planners.

**Goal Reasoning.** PlanSys2 requires users to write a Mission Controller, which serves several functions, including defining the current goal to pursue and managing the Executor component. Their tutorial suggests that users harness this mechanism to implement a goal reasoning mechanism. This may allow users to better adapt to certain situations, but requires understanding and implementing such a mechanism manually.

**Pre-generated Plans and Decompositions.** PlanSys2 supports defining Behavior Trees (BT) using XML, which includes logic to activate robot skills in a predefined plan to achieve a combined effect. Each BT is represented as a single PDDL action. It is important to note that these BTs are distinct from those autogenerated by the Executor.

**Monitoring and Sensing.** The robot skills can update the world state, influencing the success or failure of PDDL actions based on concurrent PDDL action-conditions. Therefore, it's possible to manually implement *synchronous sensing* and *monitoring*.

*3. State Representation and Update:* Plansys2 stores world state information in PDDL format within its Domain and Problem experts, which can be queried using a C++ API, ROS2 services, and a convenient shell application. The domain data is loaded from a PDDL file, while the world state can either be loaded from a problem file or updated through C++, ROS2 services, or shell APIs. Plansys2 maintains a single possible state of the world and stores this information in memory, avoiding the use of a database.

Similarly to ROSPlan, the *execution state* is not saved within the system, and it is the user's responsibility to handle translations from the *world state* to specific code parameters or to transfer *execution state* information between skills. As in ROSPlan, their interfaces allow external agents or human users to query and update the *world state*.

*4. Integration:* When integrating skill code with Plansys2, users should follow these steps: 1. **Describe a PDDL Domain File** to define the robot's possible actions and environment. 2. **Write a ROS Node for Each Robot Skill** by extending the 'ActionExecutorClient', where user code maps the grounded PDDL action parameters to low-level robot code activation. The base class handles the activation during plan execution, and Plansys2 automatically applies action effects and checks conditions. 3. **Create a Launch File** that includes the PDDL domain file and lists each of the PDDL actions along with their respective 'ActionExecutorClient' nodes responsible for execution. 4. **Implement a Mission Controller** which should (a) retrieve the domain information from the 'DomainExpertClient', (b) initialize the problem information with instances, predicate values, and the goal, (c) generate the plan using the 'PlannerClient', and (d) execute the plan using the 'ExecutorClient'. 5. (optional) **Implement Goal Reasoning Code** to determine the next goal once the current plan has finished executing.

Plansys2 is very effective for deterministic domains, allowing the modeling of carefully engineered industrial settings well. Yet, to handle non-deterministic domains, its automatic update process must be bypassed, as these require either a more involved update process and the use of replanning or some other mechanism, such as a behavior tree. Handling these cases requires non-trivial modifications.

To assist robot development, PlanSys2 provides detailed logs for various internal operations, including generated plans, specific plan execution statuses, and actions selected by the auction mechanism. This enhances system transparency, explainability, and debuggability. In addition to ROS services for querying (domain and problem files) and updating (problem files) similar to ROSPlan, PlanSys2 offers a C++ client API and a convenient shell application. It also provides DOT graph generation to visualize (temporal) plans. Moreover, plansys can consolidate multiple domains into a single unified domain, facilitating the seamless integration of various distinct projects. This feature is particularly useful for combining unrelated systems into a cohesive domain, such as a navigating mobile base robot and an arm manipulator equipped with camera-based object detection.

**Interaction.** Similar to ROSPlan but with enhanced interfaces, Plansys2 enables humans and other agents to interact with the system by modifying its world state content through the Problem Expert interfaces. These interfaces not only allow skills to update the state but also enable them to read the generated plan, facilitating the consideration of future executions when applying the skill code.

*5. Impact:* Plansys2 supports ROS2, the leading robotic framework today, making it relevant to a wide audience of researchers and industry engineers. It features a user-friendly GitHub page that offers tutorials on the installation process and common use cases. A significant impact of the system is the selection of the Clips Executive designers to wrap Plansys2 within their system, as a tool for planning and execution in ROS2.

### D. SkiROS2 [42]

SkiROS2 is the advanced successor to SkiROS [43] (see also [44], [45], [46], [47], and [48]). It is a platform designed for industrial robots, emphasizing knowledge integration. The architecture of SkiROS2 consists of three main components. At its core, the **World Model (WM)** stores an ontology in OWL format, capturing both low-level (*execution state*, e.g., exact positions) and high-level (*world state*, e.g., abstract decision-making) information about the robot environment. The **Task Manager**, in charge of planning, automatically generates PDDL domain and problem files from the WM knowledge and activates a planner to generate a plan.

The **Skill Manager** is responsible for loading skills and managing their execution according to the plan, which is automatically translated into an extended behavior tree for seamless execution in ROS. SkiROS2 also supports scenarios involving multiple robots, where each robot is equipped with its own Skill Manager, yet all share the same World Model (WM). The planning model of SkiROS2 is described by an initial ontology augmented by skill models, each represented using a Python SkillDescription class.

We now review different aspects of the system.

*1. Modeling:* **Supported Planning Types:** SkiROS2 supports the PDDL action description language for both classical and temporal planning.

*2. Decision Making:* SkiROS2's decision-making process is based on a hybrid architecture. For medium-to-long-term strategies, it employs a deliberative approach using AI planning. At a lower level, it incorporates fast, reactive behaviors described using a Behavior Trees (BT).

**Planners.** SkiROS2 uses the *Temporal Fast Downward* [41] planner.

**Goal Reasoning.** Goal Reasoning (GR) falls outside the scope of SkiROS2. In its primary use case, goals are sent to the Task Manager from the Manufacturing Execution System (MES).

**Pre-generated Plans and Decompositions.** SkiROS2 users can define compound tasks using extended Behavior Trees, incorporating both primitive and compound skills. This setup enables the execution of a predefined policy to handle complex yet repetitive scenarios. These compound tasks can be triggered by the PDDL planner, integrating them into the overall planning strategy.

**Monitoring and Sensing.** SkiROS2 automatically checks the PDDL preconditions, overall conditions, and postconditions of actions against the World Model (WM). The WM acts as a digital twin of the real world, facilitating planning. Each precondition is linked to a robot precondition-check code that senses and updates the WM to ensure compliance. If the preconditions are not met, replanning is invoked.

Next, the skill is executed, and overall conditions are verified at every tick of the Behavior Tree (BT). If an overall condition is not met during skill execution, a human operator is called. After execution, a post-condition check procedure is initiated to sense and update the state. If the state does not comply with the post-condition, the skill is activated several times before a human operator is called.

Skills can update both high-level (world state) and low-level (execution state) aspects of the WM, supporting both synchronous and asynchronous sensing and monitoring of the system. However, corrective behaviors may involve a human operator in industrial settings.

*3. State Representation and Update:* In SkiROS2, the world state is stored in the World Model server using OWL (Web Ontology Language), which is based on RDF (Resource Description Framework). RDF is a standard for representing information on the web, facilitating the encoding, exchange, and reuse of structured metadata to make data machine-readable and interoperable. In robotics, ontologies like OWL organize knowledge about systems by defining elements and their relationships within a domain through Description Logic. This structured approach allows for easy updates and adaptations.

Ontologies are structured in layers: foundational layers address broad concepts, while upper layers focus on specific domain details. SkiROS2 builds upon the IEEE Standard 1872™-2015 [49] and its core ontology, CORA, enhancing it with a domain-specific model for pick and place operations.

Moreover, SkiROS2 provides Python interfaces to update the World Model (WM), which represents a single deterministic state of the environment. Although using such an ontology is a disciplined approach, it requires special attention and effort from the programmer to ensure it is consistent and comprehensive.

Interactions between skills can occur in two ways: first, by updating the execution state, which is part of the WM and accessible to all skills; and second, through the Behavior Tree (BT) blackboard, which can store a range of data from simple coordinates to large images. Humans interact with the system via its GUI, allowing them to modify the WM. In scenarios involving multiple robots, all robots share the same WM, eliminating the need for inter-robot interaction.

SkiROS2 offers a suite of tools to assist robot development and enhance its integration with ROS1 packages. Firstly, it enables seamless integration of world model elements into the ROS tf system. Secondly, SkiROS2 simplifies the conversion of ROS actions into SkiROS2 skills. Thirdly, it features a GUI implemented via ROS' rqt, allowing users to view and modify the world model and manage the activation and termination of specific skills. Additionally, SkiROS2 includes a spatial reasoner (AauSpatialReasoner) for converting or reasoning with data from the world model. Currently, SkiROS2 operates exclusively on the older ROS1 system, but development is underway to add support for ROS2.

*4. Integration:* To integrate their code with SkiROS2, users must follow a multi-step process that starts with developing an ontology that extends SkiROS2's base ontology by incorporating details about their specific robot, each hardware device, and their properties. Subsequently, users must review the configurations of the World Model, Skill Manager, Task Manager, and GUI launch files to ensure proper linkage. For example, the Skill Manager must be associated with a specific robot ontology definition, which is linked within the Task Manager launch file.

The next phase involves the creation of skills. Users must prepare and modify the necessary ROS files. Additionally, they need to develop a Skill model for each skill by extending the SkillDescription Python class. This class includes callback functions triggered at various stages such as loading, preemption, pre-start, execution, and post-execution. Users can insert code within each callback to manipulate both the robot and the WM data (*world* and *execution state*). Within this class, users are required to specify the input and output parameters of the skills, the conditions under which they operate, and implement the execute function that activates the robot itself. Any modifications made to the World Model by the skills must be incorporated within this class.

*5. Impact:* The system requires users to follow strict guidelines and demands significant integration effort. Nev-

ertheless, it offers comprehensive and user-friendly documentation and tutorials, making it easier for new users to understand and use the system.

SkiROS2 is a well-designed system equipped with a comprehensive set of tools, making it an excellent first choice for programming robots to perform deterministic tasks similar to those in some industrial settings. However, its robust and detailed design requires strict adherence from programmers, especially in writing the ontology, and it does not model partial observability and noisy sensing, as well as probabilistic skill outcomes, which may be less common in industrial settings.

*E. Skill Based Architecture [3] [50] [51] [52] [53]*

The skill-based architecture [53] places great emphasis on supporting safe and reliable behavior. The system's methodology and tools provide guarantees about the robot's behavior, supporting automated model checking and fault-tolerant design.

The system is designed to abstract functional code into interrelated skill sets, each divided into skills. It features a *decision layer* containing a *Mission Controller*, which interacts with the *SkillSet Interface* to schedule skills. Each skill set is managed by a single *Manager*, responsible for overseeing all skills and system data.

The *Mission Controller*, responsible for accomplishing the mission, can be implemented using either an FSM, behavior trees (BT), Petri-nets [52], or a PDDL solver.

We now review different aspects of the system.

*1. Modeling:* The abstraction of functional code is done using a custom *robot-language* to model the model. The *robot-language* describes *skills*, which are grouped into *skill sets* of interrelated skills (e.g., one for robot movement skills and another for communication skills), each with its own decision layer. Each *skill set* has *Resources*, which act as state variables (defining the *world state*) for decision making. Each *Resource* is associated with a state machine that determines possible values and legal transitions. *Skills* have start, successful end, interrupt, and failure effects, defined by their effects on Resources. Skill termination is defined by postconditions. Effects and conditions relate to Resource values, and effects can only occur if the Resource's current state and effect value have a transition in the Resource FSM.

**Supported Planning Types:** The *robot-language* model can be automatically translated into PDDL to solve *classical* and *temporal planning* problems, but the system cannot reason about probabilistic outcomes or uncertainty in planning.

*2. Decision Making:* **Planners.** In [3] the OPTIC classical temporal planner (see [54]) was used

**Goal Reasoning.** Not supported.

**Pre-generated Plans and Decompositions.** The system's division into skill sets can be seen as dividing the robot task into several agents, each with a different sub-task. Additionally, predefined policies are supported when using behavior trees (BTs) to define the decision layer, yet they are not supported when using a solver (for specifying a compound skill, for example).

**Monitoring and Sensing.** The system has event mechanisms that allow skills to update the Resources used for planning, supporting asynchronous sensing. However, its PDDL-solver plan generation and execution are not automatic, so monitoring is not enabled when using planning. When a BT is used for control, it defines a reactive behavior that can be treated as asynchronous monitoring. Nevertheless, the focus here is on AI planning capabilities, and in this aspect, monitoring is not supported.

*3. State Representation and Update:* While the *world state* in the system is defined by the values of *Resources*, the *execution state* of the system is set by *robot language* objects called *Data* parameters, which define low-level variables used as inputs or outputs for skills. Sometimes, a *Resource* corresponds to a Data variable (e.g., a Battery Resource with Empty, Full states and a robot-battery-level float variable). The *Resources* can be updated both by definitions inside the skill set, as described in the *robot-language*, and externally using low-level code. External *Resource* value updates are performed using *events* defined in the *robot-language*, which can be triggered externally. Each *event* has a guard condition related to *Resource* values; if the condition is met, the *event* effects update the *Resource* value. This update occurs only if all updates are valid, meaning that the current *Resource* value and the effect have a valid transition defined in the *Resource's* FSM.

*Events* defined in the *robot-language SkillSet* are not only useful for skill code that updates the *Resource* state, but they can also be used (with additional code) for humans or other agents to change the robot state. Similarly, *Resource* values can be queried using the *Resource Manager*.

*4. Integration:* Users must first define the skills using the *robot-language* and group them into skill sets; for a simple robot, a single skill set can group all the skills. Next, users should perform model checking (see 'Automated model checker' below) for possible model corrections. Model checking is automated, but the fixes are manually added. Then, the system automatically generates the *SkillSet Manager* (both ROS and ROS2 are supported), responsible for the interfaces to activate each skill and the semantics for doing so. Moreover, based on the robot language specifications, the system auto-generates code with hook functions at different points in each skill lifecycle, such as the 'validate hook' that checks if the functional layer is ready to run a skill, or the 'interrupt hook' that handles the case when a skill is stopped. Users must write code within these hook functions in the *SkillSet Manager* to connect the high-level *robot language* model to the functional layer.

Finally, users should generate a skill fault model and ensure all possible failures are detected and handled (see 'Fault-Tree Analysis' below). If not, they must repair the functional layer.

The system offers: **Automated model checker**. When writing a model, it is common to encounter logical errors that produce inconsistencies. The system includes an automated process to identify such errors. The logical consistency check involves automatically converting the *robot-language* skill sets model to *SMT* (Satisfaction Modulo

Theory) and using a solver (Z3 [55]) to identify logical inconsistencies. The verification process includes several steps: *Guards* (conditions on resources) are checked to ensure each condition has at least one possibility to be true and one to be false; otherwise, it is deemed illogical. This check applies to *events* as well as pre, post, and invariant conditions. *Event* effect checking identifies definition errors by verifying that if the guard holds true, all effects can be executed, meaning that the *Resource* state machine can transition from the current state to the state in the effect. Pre or invariant conditions are checked in groups, where the solver ensures that if all conditions hold, the inspected condition can be either false or true, making it logically meaningful. Finally, any discrepancies found should be manually corrected in the skill set definition.

**Fault-Tree Analysis**. To approve robots, both the high-level model and the functional layer need to be checked. Fault Tree Analysis (FTA) is part of this system's methodology. It involves a manual process where engineers find undetectable or unhandled errors in the model and functional layer. The system helps by auto-generating the roots of a Fault Tree (FT). The process ensures system integrity by identifying faults, resulting in recommendations to improve the skill set model or functional layer. Although manual, this methodology can lead to a more stable and robust robot.

The skill system offers a strong, sophisticated methodology with added-value tools that emphasize correctness and can help build more robust robots. The automated model checker is a significant feature that assists users without requiring manual labor. However, this system requires non-trivial manual programming from developers, as the system must be rigidly and rigorously defined. Specifically, Resources are enumerated so that every possible value and all legal transitions must be defined, which can be tedious for real-world scenarios with many values and transitions. Events must also be well-defined, and if a user wants to change how a skill updates a Resource value, they must modify the hook, the event, and possibly the Resource FSM. Moreover, the FTA process is mainly manual and requires significant user effort.

The auto-generation of PDDL is another potentially important tool for planning but the lack of support for auto-execution of the generated PDDL plans makes this feature somewhat less practical.

*5. Impact:* The system features comprehensive documentation and tutorials, including videos to explain the system concept, describe experiments with the system, and how to use it (see [56]).

### F. The Autonomous Robot Operating System (AOS [57])

The AOS is an *MPER* designed with three main goals: native support for probabilistic skill outcomes, partial observability, and noisy sensing; minimizing the integration effort; and simplifying model construction. The first goal is achieved by using partially observable Markov decision processes (POMDP) as the underlying model and a planning model that stores belief states rather than states. The second goal is achieved by (1) maintaining a clear abstraction barrier between the system and its skill code so that skills are treated as black boxes on which no assumptions are made, and hence no modification is required. (2) Providing a mapping language for declaratively specifying the relation between skill models and skill code, which the system uses to automatically generate all integration code. The third goal is achieved by allowing the use of code for describing skill models via generative models.

The system has a *Web API* server that receives HTTP requests to integrate documented code, query, and operate the robot. When an integration request with documented code is received, the system auto-generates, compiles, and executes the necessary code, along with the skills' code, completely automatically. We now review different aspects of the system.

*1. Modeling:* **Supported Planning Types:** The system supports POMDPs, allowing it to reason about probabilistic skill outcomes, partial observability of the environment, noisy sensing, and a rich goal specification described as a utility function. However, temporal and concurrent actions are not supported.

The AOS uses as its ADL the Skill Documentation Language (SDL). A single Environment File (EF) describes the possible state variables, the initial probabilistic belief state, extrinsic events in the environment that the robot does not invoke, and a utility function that specifies desired or undesired states and their associated values.

Each robot skill is associated with two files: a Skill Documentation (SD) file and an abstraction mapping (AM) file. The SD file specifies a generative model of the skill, i.e., a model that can correctly sample the next state. The advantage of this approach is that programmers can use a probabilistic programming language to specify this model, i.e., a programming language enhanced with sampling functions. (Specifically, C++ enhanced with such functions). As argued in [25], this allows for more expressive specifications that are easier to write.

The AM file describes the relationship between model parameters and skill-code parameters and between skill-code return values and observations (which are the return value of actions in POMDPs).

*2. Decision Making:* **Planners.** The AOS supports an enhanced version of POMCP [58]), capable of solving POMDPs with large and complex state spaces, including continuous state variables. The AOS uses the solution of the underlying MDP (the same POMDP model given that we know the current state) as a heuristics. The MDP heuristics can be described inside the EF file or auto-generated using Deep Reinforcement Learning (DRL).

For small domains, the AOS can automatically create a detailed POMDP model (in .POMDP file format [59]) by sampling from the generative model described in SDL. This POMDP model is then fed into SARSOP [60], a point-based solver that is efficient for small POMDP domains and creates full optimal policies offline. In these cases, the AOS switches from using the online POMCP solver to SARSOP smoothly.

| | AOS | ROSPlan | CLIPS Ex. | Skill-Based Arch. | SkiROS2 | Plansys2 |
|---|---|---|---|---|---|---|
| ADL | Structured PPL | PDDL/RDDL | PDDL | PDDL | PDDL | PDDL |
| Model Generation | X | X | X | From robot-language | Requires Ontology | X |
| Conc. & Durative Actions | X | ✓ | ✓ | ✓ | only durative | ✓ |
| Probabilistic Effects | ✓ | ✓(with RDDL) | X | X | X | X |
| Partial Observability | ✓ | X | X | X | X | X |
| Noisy Sensing | ✓ | X | X | X | X | X |
| Plug'n Play | ✓ | X | X | X | X | X |
| Supported Platforms | ROS (extendable) | ROS | ROS2 & Fawkes | ROS & ROS2 | ROS & ROS2 In Progress | ROS2 |
| Automated Model Checking | X | X | X | ✓ | X | X |
| Goal Reasoning | X | X | ✓ | X | Manually coded | X |
| Pre-gen Plans & Decomps | X | FSMs | Skill sequences for goals | Skill Set Separation | (Nested) BTs | BTs |
| Sensing | Active | Active+Passive | Active | Active | Active | Active |
| Auto Mapping | ✓ | Passive sensing only | X | X | X | X |
| Execution Monitoring | ✓ | X | ✓ | ✓ | X | X |

TABLE I: System Comparison. **Model Generation**: Planning model generated automatically from code. **Plug'n Play:** System can use a new skill given its documented code, only. **Sensing:** *Active* = sensing based on skill return values. *Passive* = sensing processes run in the background, updating the state without explicit activation. **Auto Mapping:** Structured translation of action parameters into code parameters and low-level sensor to model-level data. **Execution Monitoring:** Model-level data triggers automated plan update.

**Goal Reasoning.** The AOS does not support a Goal Reasoning (GR) mechanism. However, it does maintain the previous belief state after when new SDL files are loaded, allowing engineers to smoothly transition into a new task. Users can implement a GR mechanism by sending integration requests with different EF files that reflect new utility functions.

**Pre-generated Plans and Decompositions** The AOS does not support predefined plans or planning problem decomposition.

**Monitoring and Sensing.** The AOS receives observations (state updates) only upon skill termination. While it has a rich mechanism to define the observation value, it does not allow intermediate state updates, meaning it only supports *Synchronous Sensing*. The AOS performs planning every time a skill terminates, supporting *Synchronous Monitoring*. However, once a skill starts, there is no option to stop the skill or invoke planning, so it does not support *Asynchronous Monitoring*.

*3. State Representation and Update:* In POMDPs, the state is modeled as a distribution over potential states. In the AOS, this is managed using a particle set, where each particle represents a possible world state with specific values assigned to each state variable. This particle set serves as an approximation of the belief state outlined in the SDL model. The size of the particle set is user-configurable, and the system can manage tens of thousands of states per belief state, assuming the state size itself is not overly large.

When a robot skill code is activated, it returns an observation that is used to update the particle filter [57].

While the particle set is primarily stored in internal memory, a configurable portion can be stored in the Unifying Layer, making it accessible for queries, such as those performed by the GUI.

*4. Integration:* Beyond the skill code, AOS users need only supply *SDL* documentation files providing true plug-and-play capability, making incremental development easy.

The *AM* file contains information that enables the system to understand how to activate the skill code and how to translate the outcome of its execution into an observation.

Human interaction in the AOS is limited to loading new SDL files. Planner-skills interaction occurs solely through skill activations and the observations returned, and skills do not have access to the generated policy.

The AOS offers a variety of tools to facilitate easy robot development. Users can integrate their documentation to run initial simulations without activating the robot, allowing them to inspect their model and the generated policy. Unlike PDDL or RDDL, the planning model in AOS can be debugged. Users can manually activate different actions to observe their effects on the model, with the option to apply these actions on the robot or not.

The system also provides a GUI that enables users to perform various actions. Notably, it allows users to view the current robot belief state in graphs describing the distribution of different state variables. Users can trace the history of the robot's belief state, the actions selected, and the observations returned during or after execution. The GUI can display the belief state either as graphs or through an image visualization where an image is constructed from the possible world states; each state is represented by an image, and these are combined such that state images are faded or bold based on their likelihood.

*5. Impact:* The AOS has an elaborate GitHub [61] page with documentation and video tutorials. Currently, AOS only supports ROS1 on Ubuntu 20.04, which is becoming obsolete. However, work is underway on the next version, which will support ROS2 and include additional features.

## V. FUTURE RESEARCH DIRECTIONS

After reviewing the state-of-the-art in Model-based Planning and Execution systems for Robotics (MPER), we now describe some existing advanced features and some new ones, that we believe should be part of future research on MPERs.

### A. Multi-level Hybrid Design

AI planning is primarily used for deliberative mid to long-term operations of autonomous robots. However, this approach is less reactive compared to rule-based reactive robot architectures that act directly based on sensor readings. A common solution is hybrid systems where AI planning handles long-term decisions to schedule various skills, while the skill code itself acts as a low-level reactive layer.

Several systems incorporate an intermediate layer, enabling users to define compound skills alongside their primitive skills. These compound skills are used when complex behaviors are needed for repetitive tasks. The compound skills are predefined policies modeled as BTs or FSMs that activate primitive skills. The high-level AI planner (the deliberative layer) schedules the complex skills as actions, supporting a type of hierarchical decomposition.

Whereas in the above systems, the planner can invoke complex skills that are composed of primitive skills, the Clips Executive (CX) design introduced an innovative concept where users define a high-level plan (the goal tree) in which the leaves represent goals that are to be solved by an AI planner online. This approach utilizes the AI planner as a primitive skill within a predefined high-level plan.

This immediately suggests providing a more flexible, hybrid multi-level MPER architecture where both solvers and predefined policies are treated as (non-primitive) skills. These skills can select and trigger other skills, which may themselves be solvers, predefined policies, or primitive skills. The only limitations are that leaf skills must be primitive, and the skills under a solver need to be modeled appropriately.

An example of using such an architecture would be a robot that cleans a building. The robot must select a floor, and since changing floors takes a lot of time, it must make a well-informed decision. It receives information on the number of people and the amount of work on each floor to consider, and at the end of the day, it must be on the bottom floor.

Next, the robot needs to decide which room to clean on the selected floor. Some rooms may have people interfering with the robot, while others may require more work. The robot should aim to clean the maximum number of rooms per day. Finally, the robot must handle cleaning a single room, which involves identifying objects, picking them up, and placing each object in its proper place.

The robot's skills include navigating within and between floors, scanning a room, and picking and placing items. Such a hybrid architecture allows each task—floor selection and navigation, room selection and navigation, and cleaning a room—to be handled by either a predefined policy or a solver. The autonomous robot can be developed and tested incrementally. Robot engineers can encapsulate the entire building cleaning functionality as a skill, which may then be used by a robot that cleans a faculty campus.

While solving a single task is beneficial, such an architecture can handle the complex behaviors required for real-world scenarios. These scenarios necessitate a Goal Reasoning (GR) mechanism to determine the next tasks and their priorities. This design incorporates such a mechanism. The GR mechanism can be a predefined policy, such as in CX, or determined by an AI planner, with skill effects setting goals for lower-level solver skills. Moreover, this approach allows users to incrementally build their autonomous robot from the bottom up by solving small problems and integrating them into a broader solution or from the top down. Additionally, a predefined BT policy can activate more than one solver skill in parallel, enabling multi-agent-like solutions within the robot.

For example, an autonomous drone navigating above a potentially interesting wildlife area aims to capture video of notable events. The drone's initialization process is defined by a Behavior Tree (BT) at the root, which activates two solver skills in parallel. One solver skill is responsible for navigating the drone over the most promising areas, while the other manages the camera, deciding where and when to record. Each solver skill can utilize a combination of predefined policies and other solver skills to achieve their sub-goals.

We believe that a well-defined, user-friendly multi-level hybrid MPER architecture, leveraging these concepts, will provide maximum flexibility while remaining simple for straightforward tasks. For example, a robot with a single solver skill can efficiently schedule primitive skills.

### B. Sensing and Monitoring

Sensing and monitoring are crucial elements for building a reliable autonomous robot. ROSPlan's Sensing interface nicely support their definition. It launches a background process based on a configuration file that includes Python code. This interface can translate ROS topic messages into state variable updates and periodically call ROS services that perform more complex processing on sensor data.

For monitoring, we suggest two key improvements. First, proposed in [62], is to allow a configurable definition of a monitoring process for each skill. The monitor can be described using a BT, an FSM, or plain code. Its goal is to decide when to intervene by stopping a skill (or other components) or when to send reports to its higher authority. Second, we suggest implementing a general background safety-monitoring process. Ideally, a configuration file will specify various constraints that must be enforced (deterministically or probabilistically) at all times. For example, we may want to enforce speed limits on the robot and/or its arm velocity in the vicinity of humans or other obstacles. This mechanism can return control to the scheduling component or perform protective actions itself, such as switching to a less aggressive skill. While this can be done within the navigation or manipulation skill itself, it would require code modification. An external monitors could obviate this need. Such a mechanism can help establish general safety boundaries not limited to a specific skill. For example, ensuring the robot goes to the charging post when the battery level is low, even if the AI planner wants to take risks.

### C. Automated Planning Model Simplification

In principle, we would like to be able to provide accurate skill models. This requires expressive ADLs. However, the more complex the model, the more difficult the planning or optimization problem, in practice, and often in theory. Some systems provide some flexibility by supporting multiple models. However, this is not a unified approach.

We suggest using an ADL that is very expressive, but in which it is easy to write simpler models and annotate them accordingly or detect their properties automatically. Complementing this would be algorithms that can exploit such annotations. For example, it is desirable to be able to annotate certain skills or variables as having deterministic dynamics, or as being fully observable. Furthermore, algorithms that can automatically simplify the model to have such properties could be developed, and the simplified models could be used to provide either heuristics or faster response times when needed.

### D. Essential Features for Effective Human Control and Monitoring

The ability of humans to conveniently control and monitor a system is crucial for an MPER to ensure easy development and operation by engineers. To this effect, MPERs should move towards an 'all-in-one' interface, preferably a GUI dashboard. This dashboard should comfortably display the current world state, the currently running skills and their status, the history of skill execution and results, and the generated policy (or its approximation). This interface should also allow users to change the world state or goals, debug the model, and inspect the generated policy, preferably with the option to do so without activating the robot. Developers and users must understand the internal processes in the MPER and have some control or guidance over them. This is not just a nice-to-have feature but an essential building block of any system. While some aspects of such an interface exist, no system yet provides a fully satisfactory interface of this kind.

### E. Additional Services

Two existing useful tools in some systems are the *automated model checker* and automated translation from the system's internal representation to PDDL. We believe the following additional tools would be highly beneficial:

*1) Model Learning/Adaptation Component:* The skill model provided to an MPER is conceptualized by the skill programmer or some user. Improving it to take into account contexts not previously considered, to adapt it to environment changes or a new robotic platform are adjustments that could be automatically taken care of using a suitable learning algorithm based on the skill execution history. We believe this is a crucial component for making MPERs widely applicable.

*2) Anomaly Detection Mechanism:* It is difficult to foresee all possible use cases of a skill. Detecting unexpected situations as well as (unfortunately, likely) malicious attaches is very important for protecting the robot and its environment. Libraries for anomaly detection have already been developed, and their integration into an MPER can be quite valuable. Users can define backup procedures based on detected anomalies, improving system robustness.

*3) LLM Integration:* Several works have focused on translating natural text into planning domains (see [63], [64] and [65]). This technique can support an extremely valuable tool for MPERs that enables users to specify the model or change the task of an existing model using natural language. With such a tool, robot users could specify new tasks, constraints, or information in real-time, quickly impacting the robot's behavior. Additionally, LLMs can provide valuable explanations of the robot's current belief state or describe the generated plan or policy in a more human-friendly manner.

## VI. Summary

Model-based planning and execution architectures for robotics offer many potential advantages: greater transparency and flexibility, lower long-term development cost, and the potential for high-level performance. In this survey, we attempted to explain their main elements, the current state of this technology, and various challenges for their development. We hope this will help potential MPER users which systems suits them best, and future MPER developers improve their design and prove their value in robot programming.

## Acknowledgments

This work was supported by ISF Grant 1651/19, the Helmsley Charitable Trust through the Agricultural, Biological and Cognitive Robotics Initiative, by the Marcus Endowment Fund, and by the Lynn and William Frankel Center for Computer Science at Ben-Gurion University of the Negev. The authors extend their gratitude to Volker Krueger and Matthias Mayr for their insights regarding SkiROS, Till Hofmann for his guidance on CLIPS Ex., Charles Lesire for his advice on the skill-based architecture, and Francisco Martín Rico for his assistance with PlanSys2.

## References

[1] R. E. Fikes and N. J. Nilsson, "Strips: A new approach to the application of theorem proving to problem solving," *Artificial intelligence*, vol. 2, no. 3-4, pp. 189–208, 1971.

[2] M. Cashmore, M. Fox, D. Long, D. Magazzeni, B. Ridder, A. Carrera, N. Palomeras, N. Hurtos, and M. Carreras, "Rosplan: Planning in the robot operating system," in *ICAPS*, 2015.

[3] C. Lesire, D. Doose, and C. Grand, "Formalization of robot skills with descriptive and operational models," in *IROS*. IEEE, 2020.

[4] M. Ghallab, D. Nau, and P. Traverso, *Automated Planning*. Morgan Kaufmann, 2004.

[5] M. L. Puterman, *Markov Decision Processes: Discrete Stochastic Dynamic Programming*. Wiley, 2005.

[6] L. Guan, K. Valmeekam, S. Sreedharan, and S. Kambhampati, "Leveraging pre-trained large language models to construct and utilize world models for model-based task planning," *Advances in Neural Information Processing Systems*, vol. 36, pp. 79 081–79 094, 2023.

[7] Anonymous, "Llms can't plan, but can help planning in llm-modulo frameworks," *arXiv preprint arXiv:2311.13567*, 2023. [Online]. Available: https://arxiv.org/abs/2311.13567

[8] ——, "Evaluation of pretrained large language models in embodied planning," *arXiv preprint arXiv:2311.13217*, 2023. [Online]. Available: https://arxiv.org/abs/2311.13217

[9] V. Pallagani, K. Murugesan, B. Srivastava, F. Rossi, and L. Horesh, "Harnessing large language models for planning: A lab on strategies for success and mitigation of pitfalls," in *AAAI 2024*. IBM Research, 2024. [Online]. Available: https://research.ibm.com/publications/harnessing-large-language-models-for-planning

[10] V. Pallagani, K. Roy, B. Muppasani, F. Fabiano, A. Loreggia, K. Murugesan, B. Srivastava, F. Rossi, L. Horesh, and A. Sheth, "On the prospects of incorporating large language models (llms) in automated planning and scheduling (aps)," *arXiv preprint arXiv:2401.02500*, 2024. [Online]. Available: https://arxiv.org/abs/2401.02500

[11] Anonymous, "Towards more likely models for ai planning," *arXiv preprint arXiv:2311.13720*, 2023. [Online]. Available: https://arxiv.org/abs/2311.13720

[12] F. A. Saad, A. Lew, and V. K. Mansinghka, "Gentl: The design and implementation of probabilistic programming languages," in *Proceedings of the 40th ACM SIGPLAN Conference on Programming Language Design and Implementation (PLDI '19)*. ACM, 2015. [Online]. Available: https://opengen.github.io/gentl-docs/latest/

[13] D. W. Aha, "Goal reasoning: Foundations, emerging applications, and prospects," *AI Magazine*, vol. 39, no. 2, pp. 3–24, 2018.

[14] C. A. Knoblock, "A theory of abstraction for hierarchical planning," in *Change of Representation and Inductive Bias*. Springer, 1990, pp. 81–104.

[15] R. S. Sutton, D. Precup, and S. Singh, "Between mdps and semi-mdps: A framework for temporal abstraction in reinforcement learning," *Artificial intelligence*, vol. 112, no. 1-2, pp. 181–211, 1999.

[16] J. Stuart, "Russell & peter norvig: Artificial intelligence: A modern approach," *Prentice-Hall*, 2003.

[17] G. Canal, M. Cashmore, S. Krivić, G. Alenyà, D. Magazzeni, and C. Torras, "Probabilistic planning for robotics with rosplan," in *Annual Conference Towards Autonomous Robotic Systems*. Springer, 2019, pp. 236–250.

[18] M. Cashmore, M. Fox, D. Long, D. Magazzeni, and B. Ridder, "Rosplan: A framework for planning in ros - documentation," https://kcl-planning.github.io/ROSPlan/, 2024, accessed: 2024-07-31.

[19] D. S. S. Miranda, L. E. de Souza, and G. S. Bastos, "A rosplan-based multi-robot navigation system," in *2018 Latin American Robotic Symposium, 2018 Brazilian Symposium on Robotics (SBR) and 2018 Workshop on Robotics in Education (WRE)*. IEEE, 2018, pp. 248–253.

[20] B. Hoteit, A. Abdallah, A. Faour, I. A. Awada, A. Sorici, and A. M. Florea, "Ai planning and reasoning for a social assistive robot." *International Association for Development of the Information Society*, 2020.

[21] B. Hoteit, I. A. Awada, A. Sorici, and A. M. Florea, "Continuous operations and fully autonomy of a social service robotic system," in *2021 23rd International Symposium on Symbolic and Numeric Algorithms for Scientific Computing (SYNASC)*. IEEE, 2021, pp. 129–134.

[22] R. Borgo, M. Cashmore, and D. Magazzeni, "Towards providing explanations for ai planner decisions," *arXiv preprint arXiv:1810.06338*, 2018.

[23] D. Escudero-Rodrigo and R. Alquezar, "Study of the anchoring problem in generalist robots based on rosplan," in *Artificial Intelligence Research and Development*. IOS Press, 2016, pp. 45–50.

[24] J. Fillan, "Autonomous inspection and maintenance missions with ai planning and the rosplan framework," Master's thesis, NTNU, 2023.

[25] R. I. Brafman, D. Tolpin, and O. Wertheim, "Probabilistic programs as an action description language," in *AAAI'23*, 2023.

[26] I. D. Doychev, T. Viehmann, T. Hofmann, G. Lakemeyer, and S. Trimpe, "Goal reasoning with the clips executive in ros2," *Bachelor's Thesis*, 2021.

[27] T. Hofmann, T. Viehmann, M. Gomaa, D. Habering, T. Niemueller, G. Lakemeyer, and C. Team, "Multi-agent goal reasoning with the clips executive in the robocup logistics league." in *ICAART (1)*, 2021, pp. 80–91.

[28] T. Viehmann, N. Limpert, T. Hofmann, M. Henning, A. Ferrein, and G. Lakemeyer, "Winning the robocup logistics league with visual servoing and centralized goal reasoning," in *Robot World Cup*. Springer, 2022, pp. 300–312.

[29] T. Niemueller, G. Lakemeyer, F. Leofante, and E. Ábrahám, "Towards clips-based task execution and monitoring with smt-based decision optimization," *Proc. of PlanRob@ ICAPS*, vol. 17, 2017.

[30] T. Niemueller, T. Hofmann, and G. Lakemeyer, "Goal reasoning in the clips executive for integrated planning and execution," in *Proceedings of the International Conference on Automated Planning and Scheduling*, vol. 29, 2019, pp. 754–763.

[31] T. Niemueller, A. Ferrein, D. Beck, and G. Lakemeyer, "Design principles of the component-based robot software framework fawkes," in *Simulation, Modeling, and Programming for Autonomous Robots: Second International Conference, SIMPAR 2010, Darmstadt, Germany, November 15-18, 2010. Proceedings 2*. Springer, 2010, pp. 300–311.

[32] J. Hoffmann, "FF: The fast-forward planning system," *AI magazine*, vol. 22, no. 3, p. 57, 2001.

[33] M. Helmert, "The fast downward planning system," *Journal of Artificial Intelligence Research*, vol. 26, pp. 191–246, 2006.

[34] F. Martín, J. G. Clavero, V. Matellán, and F. J. Rodríguez, "Plansys2: A planning system framework for ros2," in *2021 IEEE/RSJ International Conference on Intelligent Robots and Systems (IROS)*. IEEE, 2021, pp. 9742–9749.

[35] A. Coles, A. Coles, M. Fox, and D. Long, "Forward-chaining partial-order planning," in *Proceedings of the International Conference on Automated Planning and Scheduling*, vol. 20, 2010, pp. 42–49.

[36] P. Eyerich, R. Mattmüller, and G. Röger, "Using the context-enhanced additive heuristic for temporal and numeric planning," in *Towards Service Robots for Everyday Environments: Recent Advances in Designing Service Robots for Complex Tasks in Everyday Environments*. Springer, 2012, pp. 49–64.

[37] T. Niemueller, T. Hofmann, and G. Lakemeyer, "Clips-based execution for pddl planners," in *ICAPS Workshop on Integrated Planning, Acting and Execution (IntEx)*, 2018.

[38] A. Ferrein and G. Lakemeyer, "Winning the robocup logistics league with visual servoing and centralized goal reasoning," *RoboCup 2022:: Robot World Cup XXV*, vol. 13561, p. 300, 2023.

Fawkes Robotics, "Fawkes," 2024, accessed: 2024-06-05. [Online]. Available: https://github.com/fawkesrobotics/fawkes

[40] ——, "Ros2 clips executive," 2024, accessed: 2024-06-05. [Online]. Available: https://github.com/fawkesrobotics/ros2-clips-executive

[41] P. Eyerich and R. M. G. Röger, "Temporal fast downward," *Department of Computer Science University of Freiburg, Germany*, 2009.

[42] M. Mayr, F. Rovida, and V. Krueger, "Skiros2: A skill-based robot control platform for ros," in *2023 IEEE/RSJ International Conference on Intelligent Robots and Systems (IROS)*. IEEE, 2023, pp. 6273–6280.

[43] F. Rovida, M. Crosby, D. Holz, A. S. Polydoros, B. Großmann, R. Petrick, and V. Krüger, "Skiros—a skill-based robot control platform on top of ros," in *Robot operating system (ROS)*, 2017.

[44] F. Rovida and V. Krüger, "Design and development of a software architecture for autonomous mobile manipulators in industrial environments," in *2015 IEEE International Conference on Industrial Technology (ICIT)*. IEEE, 2015, pp. 3288–3295.

[45] M. Crosby, R. Petrick, F. Rovida, and V. Krueger, "Integrating mission and task planning in an industrial robotics framework," in *Proceedings of the International Conference on Automated Planning and Scheduling*, vol. 27, 2017, pp. 471–479.

[46] M. Crosby, F. Rovida, M. R. Pedersen, R. P. Petrick, and V. Krüger, "Planning for robots with skills," in *4th ICAPS Workshop on Planning and Robotics 2016*. ICAPS, 2016, pp. 49–57.

[47] D. Wuthier, F. Rovida, M. Fumagalli, and V. Krüger, "Productive multitasking for industrial robots," in *2021 IEEE International Conference on Robotics and Automation (ICRA)*. IEEE, 2021, pp. 12 654–12 661.

[48] V. Krueger, F. Rovida, B. Grossmann, R. Petrick, M. Crosby, A. Charzoule, G. M. Garcia, S. Behnke, C. Toscano, and G. Veiga, "Testing the vertical and cyber-physical integration of cognitive robots in manufacturing," *Robotics and computer-integrated manufacturing*, vol. 57, pp. 213–229, 2019.

[49] I. Robotics and A. Society, "Ieee standard ontologies for robotics and automation," *IEEE Stan.*, vol. 1872, pp. 1–60, 2015.

[50] A. Albore, D. Doose, C. Grand, C. Lesire, and A. Manecy, "Skill-based architecture development for online mission reconfiguration and failure management," in *2021 IEEE/ACM 3rd International Workshop on Robotics Software Engineering (RoSE)*. IEEE, 2021, pp. 47–54.

[51] G. C. Medina, J. Guiochet, C. Lesire, and A. Manecy, "A skill fault model for autonomous systems," in *Proceedings of the 4th International Workshop on Robotics Software Engineering*, 2022, pp. 55–62.

[52] B. Pelletier, C. Lesire, D. Doose, K. Godary-Dejean, and C. Dramé-Maigné, "Skinet, a petri net generation tool for the verification of skillset-based autonomous systems," in *EPTCS 2022-Electronic Proceedings in Theoretical Computer Science*, vol. 371, 2022, pp. 120–138.

[53] A. Albore, D. Doose, C. Grand, J. Guiochet, C. Lesire, and A. Manecy, "Skill-based design of dependable robotic architectures," *RAS*, 2023.

[54] J. Benton, A. Coles, and A. Coles, "Temporal planning with preferences and time-dependent continuous costs," in *Proceedings of the International Conference on Automated Planning and Scheduling*, vol. 22, 2012, pp. 2–10.

[55] L. De Moura and N. Bjørner, "Z3: An efficient smt solver," in *International conference on Tools and Algorithms for the Construction and Analysis of Systems*. Springer, 2008, pp. 337–340.

[56] ONERA Robot Skills, "Onera skill-based robot architecture: Documentation and tutorials," 2024, accessed: 2024-08-01. [Online]. Available: https://onera-robot-skills.gitlab.io/tutorial.html

[57] O. Wertheim, D. R. Suissa, and R. I. Brafman, "Plug'n play task-level autonomy for robotics using pomdps and probabilistic programs," *IEEE Robotics and Automation Letters*, vol. 9, no. 1, pp. 587–594, 2023.

[58] D. Silver and J. Veness, "Monte-carlo planning in large pomdps," *Advances in neural information processing systems*, vol. 23, 2010.

[59] Cassandra, Anthony Rocco, "Pomdp file format specification," http://www.pomdp.org/code/pomdp-file-spec.html, 2015, accessed: July 3, 2024.

[60] H. Kurniawati, D. Hsu, and W. S. Lee, "Sarsop: Efficient point-based pomdp planning by approximating optimally reachable belief spaces," in *Proceedings of Robotics: Science and Systems IV*. The MIT Press, 2009.

[61] O. Wertheim, "Aos documentation and tutorials," 2024, accessed: 2024-08-01. [Online]. Available: https://github.com/orhaimwertheim/AOS-WebAPI/blob/master/README.md

[62] A. Leite, A. Pinto, and A. Matos, "A safety monitoring model for a faulty mobile robot," *Robotics*, vol. 7, no. 3, p. 32, 2018.

[63] P. Smirnov, F. Joublin, A. Ceravola, and M. Gienger, "Generating consistent pddl domains with large language models," *arXiv preprint arXiv:2404.07751*, 2024.

[64] Y. Xie, C. Yu, T. Zhu, J. Bai, Z. Gong, and H. Soh, "Translating natural language to planning goals with large-language models," *arXiv preprint arXiv:2302.05128*, 2023.

[65] S. Miglani and N. Yorke-Smith, "Nltopddl: One-shot learning of pddl models from natural language process manuals," in *ICAPS'20 Workshop on Knowledge Engineering for Planning and Scheduling (KEPS'20)*. ICAPS, 2020.